\pdfoutput=1
\documentclass[11pt]{article}
\usepackage[final]{acl}

\usepackage{times}
\usepackage{latexsym}
\usepackage[T1]{fontenc}
\usepackage[utf8]{inputenc}
\usepackage{microtype}
\usepackage{booktabs}
\usepackage{amsmath}
\usepackage{graphicx}
\usepackage{xcolor}
\usepackage{url}
\usepackage{pgfplots}
\pgfplotsset{compat=1.17}
\usepgfplotslibrary{groupplots}
\usetikzlibrary{positioning}

\definecolor{notecolor}{gray}{0.45}

\title{Towards Green Text Generation: Memory-Based Speculative Decoding}

\author{Antal van den Bosch \\
  Utrecht University \\
  \texttt{a.p.j.vandenbosch@uu.nl}}

\begin{document}
\maketitle

\begin{abstract}
Speculative decoding accelerates text generation with large language models by letting a cheap draft model propose tokens
that a large verifier checks in a single forward pass. We introduce Olifant, a draft
model built on a memory-based classifier. It predicts the next token from a
four-token window in under 0.3\,ms on CPU, requiring no GPU, no backpropagation, and no
accelerator memory. We measure throughput and energy under two verifier regimes on
consumer hardware. Without a KV cache, drafting is nearly free: a domain-matched Olifant
model reaches 38.1\% acceptance on EU legal text, delivering 2.43$\times$
throughput and 2.51$\times$ lower energy per token than the neural verifier model alone. With a KV-cached verifier,
itself a $\sim$10$\times$ energy saver, drafting costs matter: a TinyLlama draft
model halves throughput and triples energy despite high acceptance, whereas Olifant still yields 1.25$\times$ throughput and the lowest energy
in the study (93\,J per 100 tokens, 14$\times$ below the uncached baseline). We further show
that KV caching decouples time from energy: verification of longer drafts raises power draw,
shifting the energy-optimal draft length below the speed-optimal one. Speculative decoding
is not dead in the age of fast verifiers, provided drafting is nearly free.
\end{abstract}

\section{Introduction}
\label{sec:intro}

Text generation with large language models is sequential by nature: each token must be generated before the next,
creating a bottleneck that limits throughput even on modern hardware. \emph{Speculative
decoding}~\citep{leviathan2023speculative,chen2023speculative} addresses this by pairing a
cheap \emph{draft model} with an expensive \emph{verifier}. The draft model proposes $k$
tokens into the future; the verifier processes all $k$+1 positions in one forward pass, accepting the longest
prefix consistent with its own greedy predictions. When acceptance rates are high, many tokens
are produced per verifier call, yielding wall-clock speedups.

Speed, however, is only half of the efficiency story. Local text generation is increasingly
energy-bound: the energy cost of every verifier forward pass, not just its latency,
determines the footprint of on-device generation. This constraint binds hardest where generation cannot leave the device at all: medical and legal text applications routinely rule out cloud APIs for confidentiality reasons, and it is precisely these domains where we find the largest efficiency gains. We therefore evaluate draft models on both axes, throughput and energy per generated token (Section~\ref{sec:energy}), under
two verifier regimes. In a transformer, the attention keys and values computed for
already-processed tokens can be stored in a \emph{KV cache}, so that each new token only
requires computing attention for itself instead of re-encoding the entire sequence;
maintaining this cache turns generation into cheap incremental steps at the price of
holding per-token states in accelerator memory. We evaluate batch verification without a KV
cache, where full-context attention dominates and draft tokens are nearly free, and
KV-cached verification, where each draft position carries real cost. The two regimes give opposite answers to how long a draft should
be, and to which draft models are viable at all. We find that a neural draft model
that pays for itself against a slow verifier becomes a net loss against a fast one, while
memory-based drafting, at $\sim$0.2\,ms per token from CPU RAM, remains profitable in
both regimes.

The design of the draft model determines whether speculative decoding pays off. Neural draft
models from the same neural model type as the verifier achieve high acceptance by construction, but incur
substantial inference cost themselves. Statistical $n$-gram models, in contrast, are nearly free to query
but cannot generalize beyond observed sequences.
Recent retrieval-based approaches~\citep{he2023rest,yang2023inference} look up draft
continuations in a corpus, trading memory for prediction quality: REST retrieves
continuations whose recent context matches the generated suffix exactly, and
inference-with-reference copies spans from documents already present in the prompt.
Both draft only text that occurs verbatim in their source
(we return to this comparison in Section~\ref{sec:retrieval}).

We explore memory-based learning~\citep{daelemans2005mbl} as a novel design point.
Memory-based language modeling is rooted in two classic machine learning paradigms,
$k$-nearest neighbor classification and decision-tree learning, and stores its training
data in a prefix trie, a retrieval-oriented tree in which instances sharing feature
values share a path. Specifically, we use TiMBL's IGTree
algorithm~\citep{daelemans1997igtree,daelemans2010timbl},\footnote{TiMBL is open-source
software: \url{https://github.com/LanguageMachines/timbl}} which compresses a training set
of (context window $\to$ next token) instances into such a trie with levels ordered by
information gain, the most predictive context position at the top. IGTree classifiers are wrapped into a functional autoregressive decoder, codenamed Olifant \cite{bosch2025memorybasedlanguagemodelsefficient}. 

At query time a
single path is traversed from the root, matching the query's context tokens level by
level in this order, and the majority next token stored at the deepest matching node is
returned. A context never seen verbatim in training is thus still answered, from the
deepest matching prefix of its most informative positions; the remaining, less
informative positions are ignored. The trie is additionally pruned at training time:
nodes whose majority label agrees with that of their parent are not stored, typically
shrinking the trie by about 95\% while leaving lookups of stored contexts unchanged.
Unlike neural models, Olifant requires no GPU and no backpropagation. The contrast with an in-session $n$-gram
draft model (a baseline in this study) is of a different kind. The $n$-gram model is not trained at all: it
continuously indexes the current session's context, exploiting recency and burstiness of tokens,
whereas Olifant is a fully-fledged trainable language model in memory-based form, fixed at
inference time and trained offline on a corpus of choice. %

\section{System Description}
\label{sec:system}

\subsection{Olifant Draft Models}

We pre-train four Olifant models used for drafting. Each training corpus is tokenised with the \texttt{mistralai/Mistral-7B-v0.1} SentencePiece
tokeniser. Consecutive windows of width $w$=4 yield instances $(t_1,t_2,t_3,t_4) \to t_5$.
The four models are matched to each evaluation domain of
Section~\ref{sec:experiments}, plus an unmatched control.
\textbf{olifant-wikitext} is trained on the WikiText-103 training
split~\citep{merity2017wikitext} with 130 million tokens,\footnote{\url{https://huggingface.co/datasets/Salesforce/wikitext}}
whose test split supplies our general-text prompts; \textbf{olifant-pubmed} on the PubMed
Summarization training split~\citep{cohan2018pubmed} containing 536 million tokens,\footnote{\url{https://huggingface.co/datasets/ccdv/pubmed-summarization}}
matching the medical task; and \textbf{olifant-eurlex} on the EurLex training
split~\citep{chalkidis2022lexglue} of 94 million tokens,\footnote{\url{https://huggingface.co/datasets/coastalcph/lex_glue}, \texttt{eurlex} subset}
matching the legal task. The fourth variant, \textbf{olifant-fineweb}, is trained on one
shard of FineWeb-edu~\footnote{\url{https://huggingface.co/datasets/HuggingFaceFW/fineweb-edu}}
web-crawled educational text that overlaps with none of the three test sources, containing 1.089 billion tokens; it
serves as the largest model and domain-unmatched control, probing what an off-the-shelf
general model offers without domain match. The resulting instance bases occupy 145\,MB for the smallest model, eurlex, to 4.6\,GB for the largest model, fineweb, of CPU RAM.

At inference time, at each speculative step, the last $w$ context tokens are sent to Olifant,
which returns its prediction. %
Instance bases are loaded
lazily and released per strategy to save memory.

\subsection{Baselines}

We compare against three baseline models:

\paragraph{Plain.}
No draft model; the verifier is called once per token: the non-speculative
throughput baseline.

\paragraph{4-gram.}
A sliding-window frequency model over the accumulated context, including tokens
generated by the verifier in the current session. It therefore adapts online to the verifier's
output.

\paragraph{Neural.}
TinyLlama-1.1B-Chat-v1.0~\citep{zhang2024tinyllama}\footnote{\url{https://huggingface.co/TinyLlama/TinyLlama-1.1B-Chat-v1.0}}
via HuggingFace Transformers.
The neural draft is hosted on the MPS accelerator alongside the verifier and is
evaluated in both full sweeps.

\subsection{Draft Model Latency}

We measure standalone next-token prediction latency with 50 single-token queries
(10 prompts $\times$ 5 repetitions, no verifier involved): the 4-gram hash lookup answers
in 0.09\,ms, the Olifant variants in 0.16--0.25\,ms, %
and TinyLlama-1.1B in 254\,ms per token on CPU.

In other words, the 4-gram model and all Olifant variants predict the next token in under 0.3\,ms,
negligible against any verifier forward pass. TinyLlama needs a 1.1B-parameter forward
pass per token; we host it on the accelerator alongside the verifier, costing
$\sim$2\,GB of accelerator memory.

\subsection{Verifier and Acceptance}
\label{sec:verifier}

Mistral-7B-v0.1 base~\citep{mistral2023}\footnote{\url{https://huggingface.co/mistralai/Mistral-7B-v0.1}}
is loaded via HuggingFace Transformers on
a M1 Max GPU via MPS (\texttt{float16}). %
We evaluate two verification regimes.

In the regime Without KV caching, at each speculative step the full sequence (context $+$ $k$ draft tokens) is fed to the
model in a single forward pass; the logits at positions
$[\text{ctx}{-}1 \mathinner{..} \text{ctx}{+}k{-}1]$
are compared greedily against the draft tokens to determine the accepted prefix; i.e.,
$k$+1 token predictions are obtained at the cost of
one forward pass. No KV cache is maintained across steps, so full-context attention
dominates the cost of every step. 

In the KV-cached regime, the cached decoder prefills the context once and maintains the KV cache across steps. Upon
rejection, the cache is truncated to the accepted prefix. %
We made two design choices. First, the first draft token is verified against the
previous step's logits, which have already been computed.
Costs for the $k$-token verification pass only need to be incurred when the first draft token is accepted.
Second, acceptance is
decided at the token-ID level (exact greedy match), making the cached decoder
output-preserving by construction. In exact arithmetic its output equals plain greedy
decoding; under \texttt{float16} on MPS, rare shape-dependent kernel differences cause
occasional divergences (Section~\ref{sec:kvresults}).

\section{Experiments}
\label{sec:experiments}

\subsection{Setup}

We evaluate on three tasks:
(1) \textbf{General text}: 100 prompts from the WikiText-103 test split
(first 50 words of each article as prompts; 100 tokens generated).
(2) \textbf{Medical}: 100 prompts from the PubMed Summarization test split
(first 50 words of each article as prompts; 100 tokens generated).
(3) \textbf{Legal}: 100 prompts from the EurLex test split
(first 50 words of each EU legislative document as prompts; 100 tokens generated).

We report mean tokens per second (tps), acceptance rate, mean
accepted length (MAL), and ROUGE-L against a plain greedy
reference. Energy metrics are defined in Section~\ref{sec:energy}.
The main text quotes selected cells; complete acceptance and MAL results for
every strategy $\times$ task $\times$ $k$ cell in both regimes are tabulated in
Appendix~\ref{app:acc}.

Each regime sweeps the same 7 strategies $\times$ $k \in \{3,5,8\}$ $\times$ 300 prompts
= 6{,}300 generations. In the cached regime, wall-clock timestamps are recorded per
generation and system power is logged in parallel, so that throughput and energy for this
regime derive from a single run under identical conditions; uncached energy comes from a
separate 63-configuration measurement grid (30 prompts per cell,
Section~\ref{sec:energy}). All experiments are run on the same Mac Studio (Apple M1 Max,
32\,GB RAM) with the verifier on MPS.

\subsection{Results: Uncached Verification}
\label{sec:true}

\definecolor{olifantblue}{HTML}{2A78D6}
\definecolor{olifantgreen}{HTML}{008300}
\definecolor{olifantviolet}{HTML}{4A3AA7}
\definecolor{olifantred}{HTML}{E34948}
\definecolor{olifantgray}{HTML}{666666}

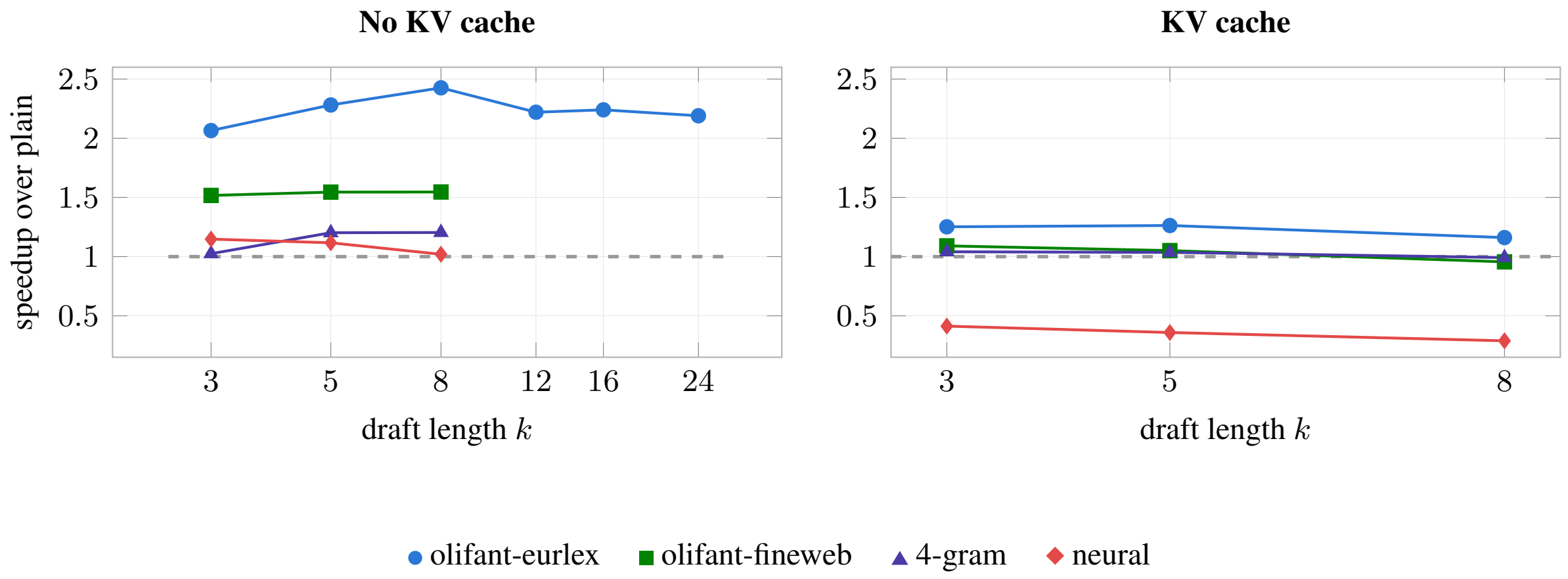
\begin{figure*}[t]
\centering
\begin{tikzpicture}
\begin{groupplot}[
    group style={group size=2 by 1, horizontal sep=1.1cm},
    width=0.52\textwidth, height=4.5cm,
    ymin=0.15, ymax=2.6,
    ytick={0.5,1.0,1.5,2.0,2.5},
    xtick=data,
    ylabel style={yshift=-2pt},
    tick label style={font=\small},
    label style={font=\small},
    title style={font=\small\bfseries},
    grid=major, grid style={gray!20, very thin},
    axis line style={gray!60},
    every axis plot/.append style={line width=1pt},
]
\nextgroupplot[title={No KV cache}, xlabel={draft length $k$},
    ylabel={speedup over plain}, xmode=log, log ticks with fixed point,
    xtick={3,5,8,12,16,24}]
\addplot[dashed, gray!80, domain=2.5:27, samples=2] {1.0};
\addplot+[thick, color=olifantblue, mark=*, mark size=1.8pt, mark options={fill=olifantblue}] coordinates { (3,2.065) (5,2.281) (8,2.426) (12,2.220) (16,2.240) (24,2.190) };
\label{plt:l:olifant-eurlex}
\addplot+[thick, color=olifantgreen, mark=square*, mark size=1.8pt, mark options={fill=olifantgreen}] coordinates { (3,1.517) (5,1.545) (8,1.546) };
\label{plt:l:olifant-fineweb}
\addplot+[thick, color=olifantviolet, mark=triangle*, mark size=1.8pt, mark options={fill=olifantviolet}] coordinates { (3,1.025) (5,1.202) (8,1.204) };
\label{plt:l:ngram}
\addplot+[thick, color=olifantred, mark=diamond*, mark size=1.8pt, mark options={fill=olifantred}] coordinates { (3,1.148) (5,1.117) (8,1.020) };
\label{plt:l:neural}
\nextgroupplot[title={KV cache}, xlabel={draft length $k$},
    xtick={3,5,8}, xmin=2.5, xmax=8.5]
\addplot[dashed, gray!80, domain=2.5:8.5, samples=2] {1.0};
\addplot+[thick, color=olifantblue, mark=*, mark size=1.8pt, mark options={fill=olifantblue}] coordinates { (3,1.252) (5,1.263) (8,1.161) };
\label{plt:r:olifant-eurlex}
\addplot+[thick, color=olifantgreen, mark=square*, mark size=1.8pt, mark options={fill=olifantgreen}] coordinates { (3,1.091) (5,1.051) (8,0.956) };
\label{plt:r:olifant-fineweb}
\addplot+[thick, color=olifantviolet, mark=triangle*, mark size=1.8pt, mark options={fill=olifantviolet}] coordinates { (3,1.041) (5,1.036) (8,0.992) };
\label{plt:r:ngram}
\addplot+[thick, color=olifantred, mark=diamond*, mark size=1.8pt, mark options={fill=olifantred}] coordinates { (3,0.413) (5,0.359) (8,0.289) };
\label{plt:r:neural}
\end{groupplot}
\node[below=0.72cm of current bounding box.south, font=\small] {
  \tikz\fill[olifantblue] (0,0) circle (2pt);~olifant-eurlex \quad
  \tikz\fill[olifantgreen] (-2pt,-2pt) rectangle (2pt,2pt);~olifant-fineweb \quad
  \tikz\fill[olifantviolet] (0,2.4pt) -- (-2.4pt,-1.6pt) -- (2.4pt,-1.6pt) -- cycle;~4-gram \quad
  \tikz\fill[olifantred] (0,2.6pt) -- (2.6pt,0) -- (0,-2.6pt) -- (-2.6pt,0) -- cycle;~neural
};
\end{tikzpicture}
\caption{Speedup over plain decoding vs.\ draft length $k$ on legal text, under
the two verifier regimes (shared $y$-axis). Without a KV cache (left), draft
tokens are nearly free and speedup grows with $k$; with a KV cache (right),
each draft position has real verification cost and every curve slopes downward:
profit requires high acceptance (domain-matched olifant-eurlex), and the
expensive neural draft falls far below break-even (dashed line).
$k \geq 12$ uncached points are from a preliminary \texttt{+D} sweep, speedups
computed against the same plain baseline.}
\label{fig:speedupk}
\end{figure*}

Table~\ref{tab:true} reports mean throughput and speedup over plain generation;
Figure~\ref{fig:speedupk} (left) plots the legal-text speedups against $k$.
Nearly every strategy achieves a genuine wall-clock speedup on every
task at every $k$ value; the single exception is olifant-pubmed on general text at $k$=3
(0.96$\times$), where the domain-mismatched model provides insufficient acceptance to offset its
overhead. The domain-matched olifant-eurlex model on legal text reaches
2.43$\times$ at $k$=8 (19.9\% acceptance) and 2.06$\times$ at $k$=3
(38.1\% acceptance), by far the best results in the study.
The domain-matched olifant-pubmed reaches 1.48--1.53$\times$ on medical text,
and olifant-wikitext is the strongest Olifant variant on its matched general-text task
(1.39--1.41$\times$), completing the domain diagonal. The unmatched control
olifant-fineweb never leads a task but is the best domain-unmatched Olifant on all
three. The 4-gram baseline benefits strongly from the no-KV-cache regime, reaching 1.42$\times$ on
general text at $k$=8.

Also, speedup increases with $k$ for olifant-eurlex on legal text:
2.06$\times$ at $k$=3, 2.28$\times$ at $k$=5, and 2.43$\times$ at $k$=8,
despite acceptance rate falling from 38.1\% to 19.9\%.
The neural draft moves in the opposite direction, from 1.15--1.20$\times$ at
$k$=3 down to 1.01--1.07$\times$ at $k$=8: the only strategy whose speedup
falls with $k$ in this regime.

\begin{table}[t]
\centering
\small
\setlength{\tabcolsep}{3pt}
\begin{tabular}{ll rrr rrr}
\toprule
 & & \multicolumn{3}{c}{tps} & \multicolumn{3}{c}{Speedup} \\
\cmidrule(lr){3-5}\cmidrule(lr){6-8}
Strategy & $k$ & Gen. & Med. & Legal & Gen. & Med. & Legal \\
\midrule
plain & --- & 2.80 & 2.71 & 2.29 & 1.00$\times$ & 1.00$\times$ & 1.00$\times$ \\
\midrule
ngram        & 3 & 3.72 & 2.74 & 2.35 & 1.33$\times$ & 1.01$\times$ & 1.03$\times$ \\
             & 5 & 3.88 & 3.40 & 2.75 & 1.39$\times$ & 1.25$\times$ & 1.20$\times$ \\
             & 8 & \textbf{3.99} & 3.47 & 2.75 & \textbf{1.42}$\times$ & 1.28$\times$ & 1.20$\times$ \\
\midrule
neural       & 3 & 3.35 & 3.16 & 2.63 & 1.20$\times$ & 1.17$\times$ & 1.15$\times$ \\
             & 5 & 3.24 & 2.99 & 2.56 & 1.16$\times$ & 1.10$\times$ & 1.12$\times$ \\
             & 8 & 2.99 & 2.74 & 2.33 & 1.07$\times$ & 1.01$\times$ & 1.02$\times$ \\
\midrule
olifant-fw   & 3 & 3.71 & 3.93 & 3.47 & 1.32$\times$ & 1.45$\times$ & 1.52$\times$ \\
             & 5 & 3.74 & 4.00 & 3.54 & 1.34$\times$ & 1.47$\times$ & 1.54$\times$ \\
             & 8 & 3.65 & 3.95 & 3.54 & 1.30$\times$ & 1.46$\times$ & 1.55$\times$ \\
\midrule
olifant-wiki & 3 & 3.94 & 3.38 & 2.72 & 1.41$\times$ & 1.25$\times$ & 1.19$\times$ \\
             & 5 & 3.90 & 3.26 & 2.53 & 1.39$\times$ & 1.20$\times$ & 1.11$\times$ \\
             & 8 & 3.90 & 3.23 & 2.42 & 1.39$\times$ & 1.19$\times$ & 1.06$\times$ \\
\midrule
olifant-pub  & 3 & 2.68 & 4.11 & 2.75 & 0.96$\times$ & 1.51$\times$ & 1.20$\times$ \\
             & 5 & 3.14 & \textbf{4.16} & 2.73 & 1.12$\times$ & \textbf{1.53}$\times$ & 1.19$\times$ \\
             & 8 & 3.03 & 4.00 & 2.64 & 1.08$\times$ & 1.48$\times$ & 1.15$\times$ \\
\midrule
olifant-eur  & 3 & 3.07 & 3.03 & 4.73 & 1.09$\times$ & 1.12$\times$ & 2.06$\times$ \\
             & 5 & 3.00 & 2.96 & 5.22 & 1.07$\times$ & 1.09$\times$ & 2.28$\times$ \\
             & 8 & 3.05 & 3.03 & \textbf{5.55} & 1.09$\times$ & 1.12$\times$ & \textbf{2.43}$\times$ \\
\bottomrule
\end{tabular}
\caption{Throughput and speedup (local Mistral-7B verifier, no KV cache),
100 prompts per task. Speedup relative to plain. olifant-fw = olifant-fineweb;
olifant-wiki = olifant-wikitext; olifant-pub = olifant-pubmed; olifant-eur = olifant-eurlex.
Bold: best per task.}
\label{tab:true}
\end{table}

Without KV caching, speedup grows with $k$ as appending $k$
draft tokens adds negligible marginal cost, so any non-zero acceptance produces a net gain,
and larger $k$ is almost always better (cf. the analysis in Section~\ref{sec:costmodel}).
Olifant-eurlex illustrates this directly: acceptance falls from 38.1\% to 19.9\% as $k$
grows from 3 to 8, yet mean accepted length grows from 1.13 to 1.53 tokens per step,
driving throughput from 4.73 to 5.55\,tps. The neural draft is the exception that
proves the rule: its $k$ draft tokens each require a sequential TinyLlama forward pass
on the accelerator, so drafting is \emph{not} free.

The gain from larger $k$ is bounded, however.
Extending the sweep to $k \in \{12, 16, 24\}$ for olifant-eurlex on legal text
(Figure~\ref{fig:speedupk}, left) shows throughput peaking at $k$=8 (2.43$\times$) and settling
near 2.2$\times$ beyond. Mean accepted length keeps growing (1.53$\to$1.62 tokens/step),
but the extra draft tokens are no longer negligible relative to the $\sim$150-token
context, making $k$=8 the practical optimum.

Output quality is unaffected. Because verification is greedy, speculative decoding
is output-preserving: an accepted
draft token is by definition the verifier's own greedy choice at that position.
Empirically, every strategy, including plain decoding, attains an identical mean
ROUGE-L of 0.956 against an independently generated greedy reference, across all tasks and
all $k$. The residual divergence from 1.0 stems from %
floating-point nondeterminism on MPS, and affects all strategies equally: the speed and
energy gains reported here cost nothing in output quality.

\subsection{Results: KV-Cached Verification}
\label{sec:kvresults}

Table~\ref{tab:kv} reports the cached-regime sweep; Figure~\ref{fig:speedupk} (right)
plots the legal-text curves on the same axis as the uncached regime.

KV caching lifts plain decoding from 2.3--2.8 to 16.3--16.9\,tps. Prompt length ceases to
matter: the 18\% uncached gap between legal and general text collapses to 3\%.
Against this faster baseline, all speculative gains are lower, and decrease with $k$,
the mirror image of the uncached regime. Domain-unmatched strategies land at break-even
(0.9--1.1$\times$): the rejection fast-path protects them from losses at $k$=3, but their
5--20\% acceptance is too low to make a positive difference. Still, 
the domain-matched olifant-eurlex retains a 
margin on legal text: 1.25$\times$ at $k$=3, and 1.26$\times$ at $k$=5 (20.6\,tps).
Acceptance
rates replicate across regimes (38.1\% on legal text in both), confirming the draft
model's behaviour is regime-independent.%

TinyLlama, a genuine 1.15--1.20$\times$ accelerator in the uncached regime, drops
to a slowdown of 0.47$\times$--0.29$\times$: each speculative step spends more time drafting
$k$ tokens sequentially than the cached verifier needs to check them.
Two-neural-model speculative decoding, the
original design, is a net loss against a fast verifier on this
hardware, while the memory-based draft model, three orders of magnitude cheaper per token,
remains profitable where its acceptance is high.

Output quality again holds: cached speculative outputs score within 1.5 ROUGE-L points
of cached plain decoding (0.955--0.959
vs.\ 0.971). ID-level greedy acceptance is output-preserving by construction; spot checks
confirm byte-identical output. The residual gap is, again, due to \texttt{float16} kernel
nondeterminism ($k$-token verification passes compute slightly different logits than $k$
single-token passes).%

\begin{table}[t]
\centering
\small
\setlength{\tabcolsep}{3pt}
\begin{tabular}{ll rrr rrr}
\toprule
 & & \multicolumn{3}{c}{tps} & \multicolumn{3}{c}{Speedup} \\
\cmidrule(lr){3-5}\cmidrule(lr){6-8}
Strategy & $k$ & Gen. & Med. & Legal & Gen. & Med. & Legal \\
\midrule
plain & --- & 16.94 & 16.84 & 16.34 & 1.00$\times$ & 1.00$\times$ & 1.00$\times$ \\
\midrule
ngram        & 3 & \textbf{18.20} & 17.72 & 17.02 & \textbf{1.07}$\times$ & 1.05$\times$ & 1.04$\times$ \\
             & 5 & 17.47 & 17.36 & 16.93 & 1.03$\times$ & 1.03$\times$ & 1.04$\times$ \\
             & 8 & 17.73 & 16.88 & 16.22 & 1.05$\times$ & 1.00$\times$ & 0.99$\times$ \\
\midrule
neural       & 3 & 7.92 & 7.51 & 6.75 & 0.47$\times$ & 0.45$\times$ & 0.41$\times$ \\
             & 5 & 6.83 & 6.33 & 5.87 & 0.40$\times$ & 0.38$\times$ & 0.36$\times$ \\
             & 8 & 5.40 & 5.00 & 4.72 & 0.32$\times$ & 0.30$\times$ & 0.29$\times$ \\
\midrule
olifant-fw   & 3 & 17.35 & 17.87 & 17.84 & 1.02$\times$ & 1.06$\times$ & 1.09$\times$ \\
             & 5 & 16.56 & 17.01 & 17.18 & 0.98$\times$ & 1.01$\times$ & 1.05$\times$ \\
             & 8 & 15.06 & 15.44 & 15.63 & 0.89$\times$ & 0.92$\times$ & 0.96$\times$ \\
\midrule
olifant-wiki & 3 & 17.84 & 17.17 & 16.64 & 1.05$\times$ & 1.02$\times$ & 1.02$\times$ \\
             & 5 & 17.14 & 16.42 & 15.94 & 1.01$\times$ & 0.98$\times$ & 0.98$\times$ \\
             & 8 & 15.62 & 15.16 & 14.78 & 0.92$\times$ & 0.90$\times$ & 0.90$\times$ \\
\midrule
olifant-pub  & 3 & 16.88 & \textbf{18.28} & 16.72 & 1.00$\times$ & \textbf{1.09}$\times$ & 1.02$\times$ \\
             & 5 & 16.50 & 17.42 & 16.04 & 0.97$\times$ & 1.03$\times$ & 0.98$\times$ \\
             & 8 & 15.59 & 15.64 & 14.99 & 0.92$\times$ & 0.93$\times$ & 0.92$\times$ \\
\midrule
olifant-eur  & 3 & 16.81 & 16.77 & 20.46 & 0.99$\times$ & 1.00$\times$ & 1.25$\times$ \\
             & 5 & 16.34 & 16.13 & \textbf{20.64} & 0.96$\times$ & 0.96$\times$ & \textbf{1.26}$\times$ \\
             & 8 & 15.47 & 15.13 & 18.98 & 0.91$\times$ & 0.90$\times$ & 1.16$\times$ \\
\bottomrule
\end{tabular}
\caption{KV-cached verification results (local Mistral-7B verifier with persistent KV cache
and rejection fast-path), 100 prompts per task. Speedup relative to cached plain.
Abbreviations as in Table~\ref{tab:true}. Bold: best tps and speedup per task.}
\label{tab:kv}
\end{table}

\section{Energy Efficiency}
\label{sec:energy}

We measure whole-system power with macOS \texttt{powermetrics} (combined CPU, GPU, and
Neural Engine power, sampled at 1\,Hz) while each strategy generates. Energy per token is
computed as $\text{J}/100\,\text{tok} = \bar{P} \cdot t_\text{wall} / n_\text{tok} \times
100$, where $\bar{P}$ is the mean power over the generation window.
For the uncached regime, each strategy $\times$ task $\times$ $k$ cell covers 30 prompts
in a dedicated measurement run. For the cached regime, energy is measured during the
throughput sweep: each generation carries a wall-clock timestamp, and per-block windows
are joined against the parallel power log, so that throughput and energy for its 100
prompts per cell come from one run under identical conditions. Background system load was minimised
during measurement; repeat measurements of the plain configuration agree within 2\% in
both regimes.

\subsection{Uncached Regime}

In the uncached regime, every draft strategy reduces energy per token on every
task at every $k$ (plain baseline: 1{,}005--1{,}326\,J per 100 tokens), as shown in Table~\ref{tab:energy}. The domain-matched Olifant models achieve the largest
savings: olifant-eurlex on legal text cuts energy per token by \textbf{2.51$\times$}
(529 vs.\ 1{,}326\,J per 100 tokens at $k$=8), and olifant-pubmed on medical text by
1.52$\times$. The neural draft is strongest on general text (727\,J, 1.40$\times$ at
$k$=5), where TinyLlama's broad coverage yields the highest acceptance, but it is 
beaten by the domain-matched Olifant models on specialised text. The domain-unmatched
olifant-fineweb still outperforms the 4-gram baseline in every single cell
(1.27--1.58$\times$ vs.\ 1.16--1.33$\times$).

\begin{table}[t]
\centering
\small
\setlength{\tabcolsep}{3pt}
\begin{tabular}{ll rrr rrr}
\toprule
 & & \multicolumn{3}{c}{J/100 tokens} & \multicolumn{3}{c}{Energy ratio} \\
\cmidrule(lr){3-5}\cmidrule(lr){6-8}
Strategy & $k$ & Gen. & Med. & Legal & Gen. & Med. & Legal \\
\midrule
plain & 3 & 1005 & 1097 & 1313 & 1.00 & 1.00 & 1.00 \\
      & 5 & 1019 & 1086 & 1326 & 1.00 & 1.00 & 1.00 \\
      & 8 & 1019 & 1095 & 1326 & 1.00 & 1.00 & 1.00 \\
\midrule
ngram & 3 & 827 & 930 & 1120 & 1.21 & 1.18 & 1.17 \\
      & 5 & 798 & 930 & 1099 & 1.28 & 1.17 & 1.21 \\
      & 8 & 767 & 945 & 1102 & 1.33 & 1.16 & 1.20 \\
\midrule
neural & 3 & 738 & 865 & 1016 & 1.36 & 1.27 & 1.29 \\
       & 5 & \textbf{727} & 854 & 989 & \textbf{1.40} & 1.27 & 1.34 \\
       & 8 & 760 & 924 & 1046 & 1.34 & 1.19 & 1.27 \\
\midrule
olifant-fw & 3 & 788 & 781 & 875 & 1.27 & 1.40 & 1.50 \\
           & 5 & 768 & 788 & 839 & 1.33 & 1.38 & 1.58 \\
           & 8 & 785 & 820 & 845 & 1.30 & 1.34 & 1.57 \\
\midrule
olifant-wiki & 3 & 815 & 916 & 1054 & 1.23 & 1.20 & 1.25 \\
             & 5 & 850 & 956 & 1064 & 1.20 & 1.14 & 1.25 \\
             & 8 & 828 & 916 & 1091 & 1.23 & 1.20 & 1.21 \\
\midrule
olifant-pub & 3 & 861 & 734 & 1081 & 1.17 & 1.49 & 1.21 \\
            & 5 & 918 & \textbf{718} & 1100 & 1.11 & 1.51 & 1.21 \\
            & 8 & 938 & 721 & 1102 & 1.09 & \textbf{1.52} & 1.20 \\
\midrule
olifant-eur & 3 & 908 & 940 & 617 & 1.11 & 1.17 & 2.13 \\
            & 5 & 919 & 961 & 550 & 1.11 & 1.13 & 2.41 \\
            & 8 & 960 & 981 & \textbf{529} & 1.06 & 1.12 & \textbf{2.51} \\
\bottomrule
\end{tabular}
\caption{Energy per generated token (J/100 tokens; lower is better) and energy ratio
vs.\ plain at the same $k$ (higher is better), measured with \texttt{powermetrics}
on Apple Silicon; 30 prompts per cell, local Mistral-7B verifier, no KV cache.
Plain rows are repeat measurements ($k$ unused). Bold: best per task.}
\label{tab:energy}
\end{table}

Across all 63 uncached configurations, mean system power stays within a narrow band
(16--31\,W, median $\approx$28\,W): no strategy meaningfully raises power draw over plain
decoding. Energy per token is therefore dominated by wall-clock time, and for the CPU-side
drafts, whose sub-millisecond lookups add no measurable power, the uncached energy
ratios track the throughput speedups of Table~\ref{tab:true} almost directly.
The neural draft deviates in the favourable direction: its energy ratios (1.19--1.40)
run ahead of its speedups (1.01--1.20), because sequential TinyLlama drafting leaves the
accelerator less saturated than the full-context verifier passes it replaces
(23.6 vs.\ 28.3\,W mean). As the next subsection shows, this
coupling of time and energy is a property of the uncached regime, not a general law.

\subsection{Cached Regime}
\label{sec:energykv}

\definecolor{olifantblue}{HTML}{2A78D6}
\definecolor{olifantgreen}{HTML}{008300}
\definecolor{olifantviolet}{HTML}{4A3AA7}
\definecolor{olifantred}{HTML}{E34948}
\definecolor{olifantgray}{HTML}{666666}

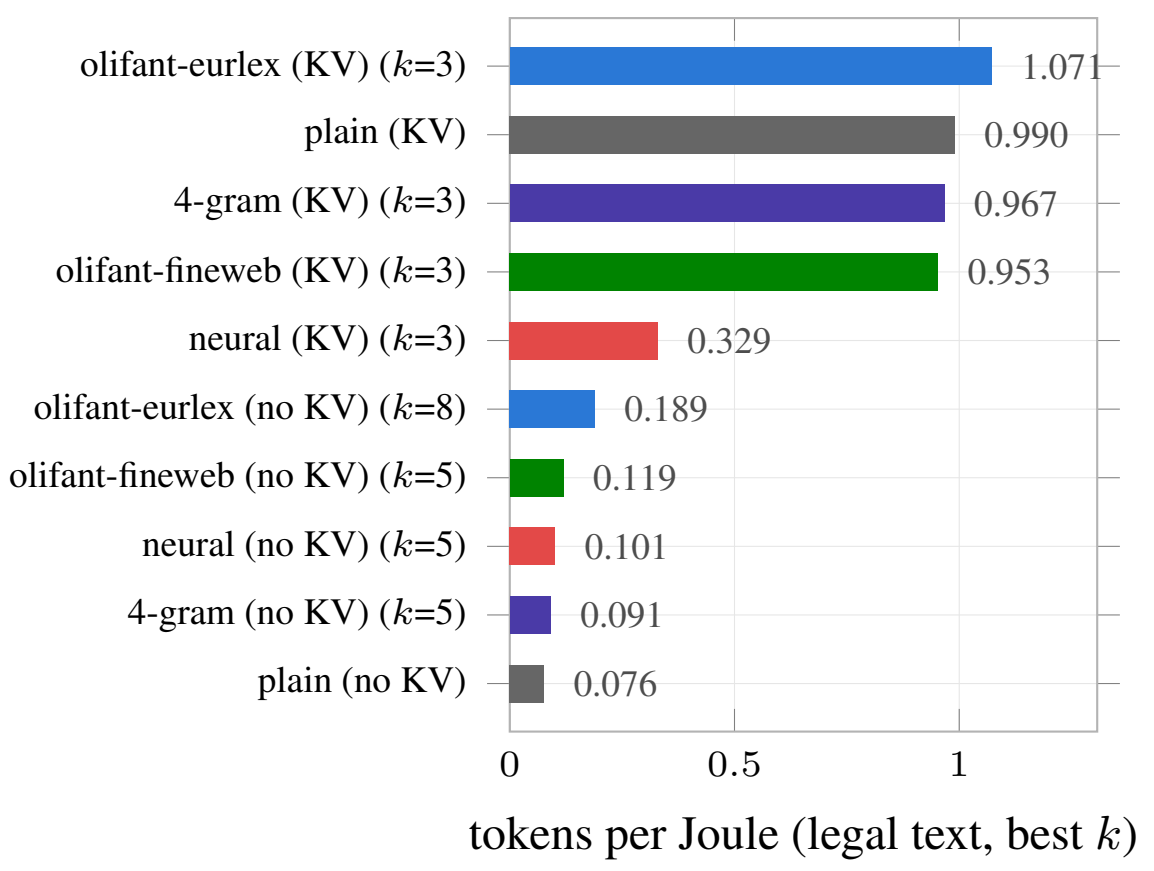
\begin{figure}[t]
\centering
\begin{tikzpicture}
\begin{axis}[
    width=0.71\columnwidth, height=6.3cm,
    xbar, xmin=0, enlarge x limits={upper, value=0.22},
    ymin=-0.7, ymax=9.7,
    clip=false,
    xticklabel style={/pgf/number format/fixed, /pgf/number format/precision=2},
    ytick={0,1,2,3,4,5,6,7,8,9},
    yticklabels={{plain (no KV)},{4-gram (no KV) ($k$=5)},{neural (no KV) ($k$=5)},{olifant-fineweb (no KV) ($k$=5)},{olifant-eurlex (no KV) ($k$=8)},{neural (KV) ($k$=3)},{olifant-fineweb (KV) ($k$=3)},{4-gram (KV) ($k$=3)},{plain (KV)},{olifant-eurlex (KV) ($k$=3)}},
    tick label style={font=\scriptsize},
    xlabel={tokens per Joule (legal text, best $k$)},
    label style={font=\small},
    axis line style={gray!60},
    grid=major, grid style={gray!20, very thin},
    y dir=normal,
]
\addplot+[xbar, bar shift=0pt, bar width=7pt, draw=none, fill=olifantgray] coordinates { (0.0761,0) };
\addplot+[xbar, bar shift=0pt, bar width=7pt, draw=none, fill=olifantviolet] coordinates { (0.0910,1) };
\addplot+[xbar, bar shift=0pt, bar width=7pt, draw=none, fill=olifantred] coordinates { (0.1011,2) };
\addplot+[xbar, bar shift=0pt, bar width=7pt, draw=none, fill=olifantgreen] coordinates { (0.1192,3) };
\addplot+[xbar, bar shift=0pt, bar width=7pt, draw=none, fill=olifantblue] coordinates { (0.1892,4) };
\addplot+[xbar, bar shift=0pt, bar width=7pt, draw=none, fill=olifantred] coordinates { (0.3295,5) };
\addplot+[xbar, bar shift=0pt, bar width=7pt, draw=none, fill=olifantgreen] coordinates { (0.9528,6) };
\addplot+[xbar, bar shift=0pt, bar width=7pt, draw=none, fill=olifantviolet] coordinates { (0.9674,7) };
\addplot+[xbar, bar shift=0pt, bar width=7pt, draw=none, fill=olifantgray] coordinates { (0.9903,8) };
\addplot+[xbar, bar shift=0pt, bar width=7pt, draw=none, fill=olifantblue] coordinates { (1.0713,9) };
\node[right, font=\scriptsize, text=black!70] at (axis cs:0.0761,0) {~0.076};
\node[right, font=\scriptsize, text=black!70] at (axis cs:0.0910,1) {~0.091};
\node[right, font=\scriptsize, text=black!70] at (axis cs:0.1011,2) {~0.101};
\node[right, font=\scriptsize, text=black!70] at (axis cs:0.1192,3) {~0.119};
\node[right, font=\scriptsize, text=black!70] at (axis cs:0.1892,4) {~0.189};
\node[right, font=\scriptsize, text=black!70] at (axis cs:0.3295,5) {~0.329};
\node[right, font=\scriptsize, text=black!70] at (axis cs:0.9528,6) {~0.953};
\node[right, font=\scriptsize, text=black!70] at (axis cs:0.9674,7) {~0.967};
\node[right, font=\scriptsize, text=black!70] at (axis cs:0.9903,8) {~0.990};
\node[right, font=\scriptsize, text=black!70] at (axis cs:1.0713,9) {~1.071};
\end{axis}
\end{tikzpicture}
\caption{Energy efficiency ladder on legal text (tokens generated per Joule,
best $k$ per strategy). KV caching is the largest single lever; domain-matched
memory-based drafting adds to it, while the neural draft costs energy in the
cached regime.}
\label{fig:ladder}
\end{figure}

Cached plain decoding costs 96--106\,J per 100 tokens (Table~\ref{tab:energykv}),
one tenth of the uncached 1{,}005--1{,}326\,J. The saving exceeds the 6--7$\times$
throughput gain, because power also drops (16.5\,W vs.\ $\approx$28\,W): one-token
steps leave the accelerator less saturated than full-context passes. Deployments
that can afford KV caching should use it; no draft model comes close to
what caching saves.

In the cached regime, power is no longer constant across strategies: mean power draw rises with
draft length, from 16.4\,W (plain) to 22.1\,W (olifant-eurlex, $k$=8), because dense
$k$-token verification passes work the accelerator harder than one-token steps. 
Consequently, time and energy deviate. 
Olifant-eurlex on legal text at $k$=8 is 1.16$\times$
faster than plain, yet spends 1.18$\times$ more energy; at $k$=3 it is both
1.25$\times$ faster and 1.09$\times$ more energy-efficient (93\,J per 100 tokens, the
study's lowest).
The
energy-optimal draft length is thus shorter than the speed-optimal one. The three
$k$-optima characterize the regimes: uncached speed favours $k$=8, cached speed
$k$=3--5, cached energy $k$=3.

TinyLlama, in contrast, adds markedly to the energy usage of the verifier: from a 1.2--1.4$\times$ energy saver in the uncached regimen to a
2.5--3.8$\times$ energy cost in the cached regimen (243--382\,J per 100 tokens). Both factors
degrade at once (doubled wall time; a second neural network occupying the accelerator),
and the deficit deepens with $k$. 

Figure~\ref{fig:ladder} visualizes the energy efficiencies of various models in terms of
tokens per Joule: cached configurations cluster an order of magnitude above uncached ones,
with cached olifant-eurlex ($k$=3, 1.07\,tok/J) at the top and the neural draft as the
only strategy that moves \emph{down} the ladder within its regime.

\begin{table}[t]
\centering
\small
\setlength{\tabcolsep}{3pt}
\begin{tabular}{ll rrr rrr}
\toprule
 & & \multicolumn{3}{c}{J/100 tokens} & \multicolumn{3}{c}{Energy ratio} \\
\cmidrule(lr){3-5}\cmidrule(lr){6-8}
Strategy & $k$ & Gen. & Med. & Legal & Gen. & Med. & Legal \\
\midrule
plain        & 3 & 96 & 97 & 102 & 1.00 & 1.00 & 1.00 \\
             & 5 & 97 & 100 & 106 & 1.00 & 1.00 & 1.00 \\
             & 8 & 96 & 97 & 101 & 1.00 & 1.00 & 1.00 \\
\midrule
ngram        & 3 & \textbf{96} & 98 & 103 & \textbf{1.00} & 0.99 & 0.99 \\
             & 5 & 122 & 108 & 110 & 0.80 & 0.92 & 0.96 \\
             & 8 & 109 & 111 & 115 & 0.88 & 0.87 & 0.88 \\
\midrule
neural       & 3 & 243 & 264 & 303 & 0.40 & 0.37 & 0.34 \\
             & 5 & 272 & 298 & 330 & 0.36 & 0.34 & 0.32 \\
             & 8 & 334 & 362 & 382 & 0.29 & 0.27 & 0.26 \\
\midrule
olifant-fw   & 3 & 109 & 103 & 105 & 0.88 & 0.94 & 0.97 \\
             & 5 & 121 & 116 & 116 & 0.81 & 0.86 & 0.92 \\
             & 8 & 136 & 132 & 132 & 0.71 & 0.73 & 0.76 \\
\midrule
olifant-wiki & 3 & 100 & 103 & 107 & 0.96 & 0.95 & 0.96 \\
             & 5 & 110 & 112 & 115 & 0.89 & 0.90 & 0.92 \\
             & 8 & 129 & 128 & 132 & 0.74 & 0.76 & 0.77 \\
\midrule
olifant-pub  & 3 & 100 & \textbf{99} & 104 & 0.96 & \textbf{0.98} & 0.98 \\
             & 5 & 106 & 111 & 113 & 0.92 & 0.90 & 0.94 \\
             & 8 & 118 & 132 & 127 & 0.81 & 0.73 & 0.79 \\
\midrule
olifant-eur  & 3 & 100 & 102 & \textbf{93} & 0.97 & 0.95 & \textbf{1.09} \\
             & 5 & 107 & 110 & 100 & 0.91 & 0.91 & 1.06 \\
             & 8 & 118 & 124 & 118 & 0.81 & 0.78 & 0.85 \\
\bottomrule
\end{tabular}
\caption{Cached-regime energy (J/100 tokens; lower is better) and energy ratio vs.\
cached plain at the same $k$, derived from the timestamped throughput sweep joined with
the 1\,Hz power log; 100 prompts per cell. Only high-acceptance domain-matched drafting
(olifant-eurlex on legal, $k \leq 5$) saves energy; ratios below 1.00 indicate an energy
cost despite any speed gain. Bold: best per task.}
\label{tab:energykv}
\end{table}

\subsection{Training Energy}
\label{sec:trainenergy}

The figures above cover inference; a green-generation claim should also count
the energy needed to build the draft models. Training an Olifant model
is a single CPU pass over the corpus: tokenisation, windowing, and IGTree trie
compilation, with no backpropagation and no accelerator.
Compiling the largest instance base (fineweb, 1.089 billion tokens) took about
six hours on the evaluation machine, roughly half a megajoule at the tens of
watts it draws for CPU-bound work: a one-time expense recouped within roughly
a thousand 100-token generations at the savings of Table~\ref{tab:energy};
the smaller corpora cost less. The neural draft model
sits at the other end of the scale: TinyLlama was pretrained on 3 trillion
tokens on 16 A100 GPUs over about 90 days~\citep{zhang2024tinyllama}, on the
order of $10^{10}$\,J counting GPU board power alone, four to five orders of
magnitude above the memory-based models. Pretrained models amortise this cost
across many downstream uses, but the contrast stands: the lifecycle energy of
a memory-based draft model is dominated by inference, not training.

\section{Discussion}
\label{sec:discussion}

\subsection{The Verifier Cost Model}
\label{sec:costmodel}

The standard break-even condition for speculative decoding is:
\begin{equation}
  \text{acc\_rate}_\text{break-even} \approx \frac{k}{k+1}
  \label{eq:breakeven}
\end{equation}
If verifying $k$ draft tokens costs as much as generating them one by one, a speculative step only pays off when nearly all of its draft tokens are accepted, and the required per-token acceptance rises with the draft length: 3 out of 4 or 75\% at $k$=3, 89\% at $k$=8.%

Without KV caching, transformer attention over the full context dominates each forward
pass, and the marginal expense of appending $k$ draft positions is negligible. The
break-even threshold collapses to near zero: any non-zero acceptance yields a speedup,
larger $k$ is better up to a plateau (Section~\ref{sec:true}), and every uncached
energy ratio exceeds $1.0$. This regime matches any deployment where the verifier
is re-invoked on the full sequence (stateless serving, or devices that cannot afford a
cache), and it is forgiving of low-acceptance drafts.

With KV caching, Equation~\ref{eq:breakeven} would predict that none of our draft models
survive: no strategy approaches 75\% acceptance. 
Two effects soften the prediction. The rejection fast-path heuristic (Section~\ref{sec:verifier}) means failed steps cost no more than plain steps, so %
a $k$-token verification pass can only achieve gains in
extra accepted tokens on steps in which the first token is accepted. 
Empirically, the resulting
threshold sits near 20--25\% first-token acceptance: below it, strategies flatline at
break-even (Table~\ref{tab:kv}); above it, which only olifant-eurlex on legal text
reaches at 38\%, a genuine margin appears. %
TinyLlama's sequential drafts cost more than the cached verification they enable,
at any acceptance.

\subsection{Speculative Decoding is Not Finished}

Against slow verification,
the original two-model speculative decoding design still performs as predicted; our uncached results reproduce this.
Verifier-side optimization (KV caching here; batching, quantization, and faster kernels
generally) shrinks the budget available for drafting, however. The field's general trajectory
away from separate draft LMs toward ever-cheaper drafting (attached heads, retrieval,
prompt lookup) concedes the point: our cached results show the separate neural draft
losing 2--3$\times$ on the same hardware where it once profited. Memory-based drafting, on the other hand,
occupies the far end of that trajectory: sub-millisecond, CPU-resident,
verifier-agnostic (any model sharing the tokeniser), and domain-adaptable by swapping an
instance base rather than retraining. It is the only draft design in this study
that remains profitable in both regimes. Speculative decoding is not finished as verifiers
get faster; drafting merely has to become nearly free, and memory-based learning is one
way to get there.

\subsection{Olifant vs.\ In-session 4-gram: Related but Distinct}

The in-session 4-gram model we compared against in this study is not a trained model: it builds its frequency table from the
accumulated context of the current session, including tokens generated by the
verifier. It therefore automatically adapts to the verifier's output distribution, and
to the recency and burstiness of words in the running text, effectively acting as an online
imitator. Olifant is the opposite design: a trained language model in memory-based form,
fixed at inference time, with no mechanism to incorporate session-level feedback. This explains why the 4-gram baseline competes well on domains where no
Olifant model was trained (Table~\ref{tab:true}): online adaptation substitutes
for static acceptance.

A second difference is that Olifant operates natively on Mistral's SentencePiece tokens, the same units the verifier
uses, so draft predictions slot directly into the verification step with no
representation mismatch. The 4-gram model predicts whitespace-delimited words, of which only the first
sub-token survives the text-fallback path; this structural mismatch caps its effective
acceptance regardless of word-level accuracy.

On domain-matched tasks, Olifant's offline training pays off decisively (38.1\% vs.\
24.7\% acceptance on legal text; 2.43$\times$ vs.\ 1.20$\times$ speedup at $k$=8). On
unmatched domains, the 4-gram's online adaptation closes most of the gap. Two extensions
suggest themselves toward the best of both approaches: training Olifant directly on the
verifier's output distribution, and augmenting the fixed instance base with a
recency/burstiness component that, like the 4-gram baseline, tracks the current session's
context. Preliminary experiments with such a hybrid are ongoing; we leave a full treatment
to future work.

\subsection{Olifant vs.\ Retrieval-Based Drafting}
\label{sec:retrieval}

Olifant is also distinct from the retrieval-based drafting methods cited in
Section~\ref{sec:intro}. REST~\citep{he2023rest} keeps the draft corpus as a
suffix-array datastore and, at each step, retrieves continuations whose recent
context matches the generated suffix exactly; inference-with-reference~\citep{yang2023inference}
copies spans from reference documents already present in the prompt, targeting
settings with high prompt-output overlap such as retrieval-augmented generation.
Both are exact-match designs: when matching, they draft long verbatim spans, and
when the current context does not occur in their source, they draft nothing.
Olifant instead compiles the corpus into a next-token predictor. The IGTree trie discards
about 95\% of the instance base at training time, its depth bounded by the
context window rather than the corpus size. Its information-gain-ordered
prefix backoff returns a prediction for every context, including contexts never
seen verbatim in training, by ignoring the least informative window positions.
The price is that Olifant drafts one token per query rather than a retrieved
span. In machine learning terms, IGTree is itself a decision-tree compression of
a $k$-nearest-neighbor learner (Section~\ref{sec:intro}), so Olifant can be read
as a decision-tree draft model with retrieval functionality. An empirical head-to-head
with REST and inference-with-reference on the same hardware and both verifier
regimes is an obvious next step (Section~\ref{sec:future}).

\subsection{Method or Text? Attributing the Domain-Match Advantage}
\label{sec:redundancy}

Olifant-eurlex's lead on legal text raises the question whether the advantage
reflects the memory-based method or merely properties of legislative text, such
as local repetition. To calibrate, we measure the redundancy of the evaluation
texts themselves. Decoding is output-preserving, so every strategy produces the
same text for a given prompt; the statistics below therefore describe the exact
texts on which all acceptance rates were computed.
Table~\ref{tab:redundancy} reports three measures over each task's 100
generations, computed on Mistral tokens to match the draft models' context
window: the \emph{within-text} repetition rate (the fraction of a text's token
4-grams that already occurred earlier in the same text), the
\emph{cross-document} overlap (the fraction of a text's distinct 4-grams that
also occur in at least one other text of the same task), and the gzip
compression ratio of the task's concatenated text as a model-free summary.

\begin{table}[t]
\centering
\small
\begin{tabular}{lrrr}
\toprule
Task & Within-text & Cross-doc. & gzip \\
\midrule
General & 29.6\% & 26.7\% & 0.29 \\
Medical & 21.2\% & 14.9\% & 0.33 \\
Legal   & 12.1\% & 73.6\% & 0.18 \\
\bottomrule
\end{tabular}
\caption{Redundancy of the evaluation texts (100 generations per task, Mistral
tokens). Within-text: mean fraction of a text's token 4-grams already seen
earlier in the same text. Cross-doc.: mean fraction of a text's distinct
4-grams that also occur in another text of the same task. gzip: compression
ratio of the task's concatenated text (lower = more redundant).}
\label{tab:redundancy}
\end{table}

Within documents, legal text is
the \emph{least} repetitive of the three tasks (12.1\% vs.\ 21.2\% medical and
29.6\% general), so within-generation repetition cannot explain the acceptance
ordering. In contrast, 73.6\% of a legal text's distinct 4-grams recur in
other legal texts, against 14.9\% and 26.7\% elsewhere, and the concatenated
legal texts compress to 0.18 of their size. This cross-document
formulaicity is what offline training on a domain corpus captures; it
is the mechanism by which memory-based drafting profits, not a confound of it.

The contribution of the method, as opposed to the text, is isolated by
comparing draft models on identical legal texts: at $k$=3, acceptance spans
7.5\% (olifant-pubmed) to 38.1\% (olifant-eurlex), with the 4-gram baseline at
24.7\% and the neural draft at 21.1\% (Appendix~\ref{app:acc}). The text is
held constant across these rows; only the draft model changes. Both factors
are thus real and separable: tasks differ in how much cross-document
redundancy there is to harvest, and draft models differ in how much of it they
capture.

\subsection{Future Work}
\label{sec:future}

Several axes are deliberately left open. First, a direct empirical comparison
with retrieval-based drafting (REST and inference-with-reference,
Section~\ref{sec:retrieval}) on the same hardware and under both verifier
regimes. Second, verifier diversity: repeating the sweeps with verifiers of
other sizes and families would test how far the cost-model analysis of
Section~\ref{sec:costmodel} carries; the same holds for server-class GPUs,
where faster verification should sharpen the cached-regime compression we
observe. Third, sampling and batching: speculative
sampling~\citep{leviathan2023speculative,chen2023speculative} extends
verification losslessly to temperature sampling, at acceptance rates that
shift with temperature, and batched serving changes the drafting budget again. Fourth,
generalization: evaluating domain-matched models on in-domain corpora from
sources other than the training collection. Fifth, the session-adaptive hybrid
trailed in Section~\ref{sec:discussion}, combining a trained instance base with
an online recency component. Finally, scaling the instance bases: eurlex, the smallest corpus in the study
at 94M tokens, produced the best draft model, leaving the scaling axis
unexplored.

\section{Conclusion}
\label{sec:conclusion}

We have presented Olifant, a speculative decoding draft model based on TiMBL memory-based
learning, and evaluated it on general, medical, and legal text against 4-gram and neural
baselines under two verifier regimes, measuring both throughput and energy on consumer
hardware.

Without a KV cache, drafting is nearly free, and nearly every strategy achieves a genuine
wall-clock speedup; only the neural draft's gain decays with draft length. The
domain-matched legal model delivers 2.43$\times$
throughput and a matching 2.51$\times$ energy saving.
With a KV cache, the baseline itself accelerates 6--7$\times$ at
one tenth the energy, and the rules change: gains diminish, the useful draft length
shrinks, and drafting cost become a prominent factor. The neural draft model loses
2--3$\times$ in both time and energy despite high acceptance, while the memory-based
draft model keeps a 1.25$\times$ positive throughput margin where its acceptance is high and reaches the lowest
energy cost in the study (93\,J per 100 tokens). The cached regime decouples time
from energy: verifying longer drafts raises power draw, so the energy-optimal draft
length sits below the speed-optimal one. Greedy verification preserves output quality
in both regimes, up to floating-point nondeterminism in the cached decoder.

In sum, verifier-side optimization keeps shrinking verification costs, so
drafting must approach zero cost to still make a difference. Memory-based drafting, at 0.2\,ms per token
from CPU RAM with no accelerator footprint, meets that bar today.

\section*{Limitations}

All experiments run on a single hardware platform (a Mac Studio\footnote{Energy was supplied by a solar panel and home battery combination, requiring almost no grid electricity.} with an Apple M1 Max chip
and 32\,GB RAM, verifier on MPS in \texttt{float16}) with a single verifier
(Mistral-7B-v0.1). On a server-class GPU, per-token verifier latency would drop further (against a higher energy footprint) and the
relative cost of drafting rises, so the cached-regime compression we measure would likely
sharpen.
Energy figures are whole-system measurements on consumer hardware with background load
minimised; they support relative comparisons on identical hardware, not absolute
cross-platform claims.
All measurements use greedy verification at batch size one; how the reported
margins shift under temperature sampling and batched serving remains to be
established (Section~\ref{sec:future}). For the domain-matched models,
evaluation prompts come from held-out test splits of the same collections used
for training (Section~\ref{sec:system}); transfer to in-domain text from other
sources is untested.

All datasets and pretrained models are available from the Hugging Face hub
(footnotes at first mention), and TiMBL and Olifant are open-source. Our
implementation and experimental scripts are released at
\url{https://github.com/antalvdb/olifant-speculative}, and the four trained
instance bases at
\url{https://huggingface.co/antalvdb/olifant-speculative-ibases}.

\bibliography{olifant}

\appendix

\section{Complete Acceptance and MAL Results}
\label{app:acc}

Tables~\ref{tab:appendix-uncached} and~\ref{tab:appendix-cached} report the
acceptance rate and mean accepted length for every draft model $\times$ task
$\times$ draft length cell, in the uncached and KV-cached verifier regimes
respectively, complementing the selected cells quoted in the main text.
Acceptance is the fraction of drafted tokens accepted by the verifier, averaged
over the 100 prompts of a task; MAL is the mean number of draft tokens accepted
per verification step. As $k$ grows, per-token acceptance declines while MAL
rises: longer drafts convert a smaller share of their tokens, but still yield
more accepted tokens per verifier call, which is why throughput can improve
with $k$ wherever drafting is cheap.

\makeatletter\setlength{\@fptop}{0pt}\makeatother
\begin{table}[htbp]
\centering
\footnotesize
\setlength{\tabcolsep}{2pt}
\begin{tabular}{lrrrrrr}
\toprule
 & \multicolumn{2}{c}{$k$=3} & \multicolumn{2}{c}{$k$=5} & \multicolumn{2}{c}{$k$=8} \\
Draft model & Acc.\% & MAL & Acc.\% & MAL & Acc.\% & MAL \\
\midrule
\multicolumn{7}{l}{\emph{General}} \\
4-gram & 22.1 & 0.31 & 15.7 & 0.35 & 11.0 & 0.39 \\
neural & 25.2 & 0.75 & 17.5 & 0.86 & 11.9 & 0.92 \\
olifant-fineweb & 12.2 & 0.36 & 7.7 & 0.38 & 4.9 & 0.38 \\
olifant-wikitext & 14.8 & 0.44 & 9.5 & 0.47 & 6.1 & 0.47 \\
olifant-pubmed & 5.1 & 0.15 & 3.2 & 0.15 & 2.0 & 0.15 \\
olifant-eurlex & 4.8 & 0.14 & 2.9 & 0.14 & 1.9 & 0.14 \\
\midrule
\multicolumn{7}{l}{\emph{Medical}} \\
4-gram & 16.4 & 0.22 & 10.9 & 0.24 & 7.3 & 0.26 \\
neural & 23.3 & 0.69 & 15.6 & 0.77 & 10.5 & 0.81 \\
olifant-fineweb & 16.0 & 0.48 & 10.5 & 0.51 & 6.9 & 0.53 \\
olifant-wikitext & 9.0 & 0.27 & 5.6 & 0.28 & 3.6 & 0.28 \\
olifant-pubmed & 18.2 & 0.54 & 11.8 & 0.58 & 7.7 & 0.59 \\
olifant-eurlex & 5.8 & 0.17 & 3.6 & 0.17 & 2.3 & 0.18 \\
\midrule
\multicolumn{7}{l}{\emph{Legal}} \\
4-gram & 24.7 & 0.19 & 16.3 & 0.20 & 10.9 & 0.21 \\
neural & 21.1 & 0.63 & 15.0 & 0.73 & 9.9 & 0.76 \\
olifant-fineweb & 18.8 & 0.56 & 12.3 & 0.60 & 8.2 & 0.63 \\
olifant-wikitext & 8.6 & 0.25 & 5.3 & 0.26 & 3.4 & 0.26 \\
olifant-pubmed & 7.5 & 0.22 & 4.6 & 0.22 & 2.9 & 0.22 \\
olifant-eurlex & 38.1 & 1.13 & 28.3 & 1.38 & 19.9 & 1.53 \\
\bottomrule
\end{tabular}
\caption{Complete acceptance rates (\%) and mean accepted lengths (MAL, tokens per verification step) for all draft models, tasks, and draft lengths $k$, \textbf{uncached} verifier regime (100 prompts per task).}
\label{tab:appendix-uncached}
\end{table}

\begin{table}[t]
\centering
\footnotesize
\setlength{\tabcolsep}{2pt}
\begin{tabular}{lrrrrrr}
\toprule
 & \multicolumn{2}{c}{$k$=3} & \multicolumn{2}{c}{$k$=5} & \multicolumn{2}{c}{$k$=8} \\
Draft model & Acc.\% & MAL & Acc.\% & MAL & Acc.\% & MAL \\
\midrule
\multicolumn{7}{l}{\emph{General}} \\
4-gram & 22.2 & 0.31 & 15.8 & 0.36 & 11.1 & 0.40 \\
neural & 25.2 & 0.75 & 17.5 & 0.86 & 11.9 & 0.92 \\
olifant-fineweb & 12.1 & 0.36 & 7.7 & 0.38 & 4.9 & 0.38 \\
olifant-wikitext & 14.9 & 0.44 & 9.5 & 0.47 & 6.1 & 0.47 \\
olifant-pubmed & 5.1 & 0.15 & 3.2 & 0.15 & 2.0 & 0.16 \\
olifant-eurlex & 4.8 & 0.14 & 3.0 & 0.14 & 1.9 & 0.15 \\
\midrule
\multicolumn{7}{l}{\emph{Medical}} \\
4-gram & 16.4 & 0.22 & 10.9 & 0.24 & 7.3 & 0.26 \\
neural & 23.3 & 0.69 & 15.6 & 0.76 & 10.4 & 0.81 \\
olifant-fineweb & 16.0 & 0.48 & 10.5 & 0.51 & 6.8 & 0.53 \\
olifant-wikitext & 9.0 & 0.27 & 5.6 & 0.28 & 3.6 & 0.28 \\
olifant-pubmed & 18.3 & 0.54 & 11.9 & 0.58 & 7.7 & 0.59 \\
olifant-eurlex & 5.9 & 0.18 & 3.6 & 0.18 & 2.3 & 0.18 \\
\midrule
\multicolumn{7}{l}{\emph{Legal}} \\
4-gram & 24.8 & 0.19 & 16.3 & 0.20 & 10.9 & 0.21 \\
neural & 21.2 & 0.63 & 15.1 & 0.74 & 10.0 & 0.77 \\
olifant-fineweb & 18.7 & 0.56 & 12.3 & 0.60 & 8.2 & 0.63 \\
olifant-wikitext & 8.6 & 0.25 & 5.3 & 0.26 & 3.4 & 0.26 \\
olifant-pubmed & 7.5 & 0.22 & 4.6 & 0.22 & 2.9 & 0.22 \\
olifant-eurlex & 38.1 & 1.13 & 28.3 & 1.38 & 20.0 & 1.54 \\
\bottomrule
\end{tabular}
\caption{Complete acceptance rates (\%) and mean accepted lengths (MAL) for all draft models, tasks, and draft lengths $k$, \textbf{KV-cached} verifier regime (100 prompts per task).}
\label{tab:appendix-cached}
\end{table}

\end{document}